%% file: root.tex
\documentclass[letterpaper, 10 pt, conference]{ieeeconf}  

\IEEEoverridecommandlockouts                              

\include{includes}

\definecolor{fraunhoferGray}{HTML}{C7C9CB}
\definecolor{fraunhoferBlack}{HTML}{060606}
\definecolor{fraunhoferWhite}{HTML}{F5F5F5}
\definecolor{fraunhoferBlue}{HTML}{005B7F}
\definecolor{fraunhoferRed}{HTML}{BB0056}
\definecolor{fraunhoferDarkRed}{HTML}{7C154D}
\definecolor{fraunhoferGreen}{HTML}{179C7D}
\definecolor{fraunhoferOrange}{HTML}{F58220}
\definecolor{fraunhoferYellow}{HTML}{FDB913}
\definecolor{fraunhoferPurple}{HTML}{9084BD}

\title{\LARGE \bf
Synset Signset Germany:\\ A Synthetic Dataset for German Traffic Sign Recognition*
}

\definecolor{fraunhoferLightBlue}{HTML}{CCDEE5}

\author{Anne Sielemann$^{1}$, Lena Loercher$^{2}$, Max-Lion Schumacher$^{2}$, Stefan Wolf$^{3,1}$,\\ Masoud Roschani$^{1}$, Jens Ziehn$^{1}$ and Juergen Beyerer$^{1,3}$ \\
\thanks{\raggedright\normalsize \href{https://synset.de/datasets/synset-signset-ger/}{\color{fraunhoferBlue}\textbf{Download Synset Signset Germany:} \href{https://synset.de/datasets/synset-signset-ger/}{synset.de/datasets/synset-signset-ger/}}\medskip}%
\thanks{\raggedright $^{1}$Fraunhofer IOSB, 76131 Karlsruhe, Germany,\newline \texttt{\footnotesize{\{anne.sielemann, stefan.wolf, masoud.roschani, jens.ziehn\}@iosb.fraunhofer.de}}}%
\thanks{\raggedright $^{2}$Fraunhofer IPA, 70569 Stuttgart, Germany,\newline \texttt{\footnotesize{\{lena.loercher; max-lion.schumacher\}\newline@ipa.fraunhofer.de}}}%
\thanks{
\raggedright
$^{3}$Karlsruhe Institute of Technology (KIT), Vision and Fusion Laboratory (IES), 76131 Karlsruhe, Germany
}%
\thanks{\raggedright *This work was supported by the Fraunhofer Internal Programs under Grant No. PREPARE 40-02702 within the ``ML4Safety'' project, as well as funded by the German Federal Ministry for Economic Affairs and Climate Action (BMWK) within the program ``New Vehicle and System Technologies'' as part of the AVEAS research project (www.aveas.org).}
}

\begin{document}

\maketitle
\thispagestyle{fancy}
\pagestyle{fancy}

\aveasSetMargins{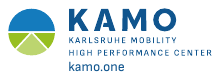}
\aveasSetIEEEFoot{2024}
\aveasSetIEEEHeadWithoutDoi{A. Sielemann et al., ``Synset Signset Germany: A Synthetic Dataset for German Traffic Sign Recognition,'' 2024 IEEE 27th International Conference on Intelligent Transportation Systems (ITSC 2024), Edmonton, Canada, 2024.}

\begin{abstract}

In this paper, we present a synthesis pipeline and dataset for training / testing data in the task of traffic sign recognition that combines the advantages of data-driven and analytical modeling: GAN-based texture generation enables data-driven dirt and wear artifacts, rendering unique and realistic traffic sign surfaces, while the analytical scene modulation achieves physically correct lighting and allows detailed parameterization. In particular, the latter opens up applications in the context of explainable AI (XAI) and robustness tests due to the possibility of evaluating the sensitivity to parameter changes, which we demonstrate with experiments. Our resulting synthetic traffic sign recognition dataset Synset Signset Germany contains a total of 105\,500 images of 211 different German traffic sign classes, including newly published (2020) and thus comparatively rare traffic signs. In addition to a mask and a segmentation image, we also provide extensive metadata including the stochastically selected environment and imaging effect parameters for each image. We evaluate the degree of realism of Synset Signset Germany on the real-world German Traffic Sign Recognition Benchmark (GTSRB) and in comparison to CATERED, a state-of-the-art synthetic traffic sign recognition dataset.
\end{abstract}

\section{INTRODUCTION}\label{sec:introduction}

\PARstart{W}{ithin} the development of machine learning (ML) and artificial intelligence (AI), and with the substantial advances achieved in the performance, particularly of deep learning, the attention of research and development has shifted to include not only maximum performance of ML and AI, but also properties relating to how this performance is achieved---namely concerning the methods for providing data of sufficient quality, recency and practical costs, and understanding the system behavior w.r.t. the real world, in terms of explainable AI (XAI), robustness, and validation. 

\begin{figure}[H]
    \centering
    \includegraphics[width=0.24\columnwidth]{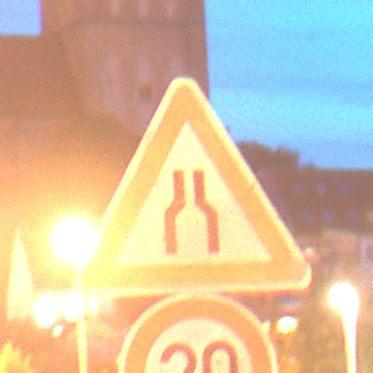}
    \hfill
    \includegraphics[width=0.24\columnwidth]{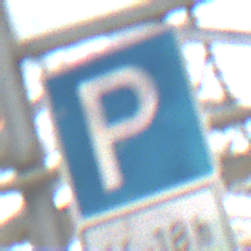}
    \hfill
    \includegraphics[width=0.24\columnwidth]{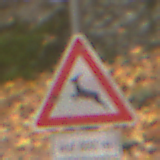}
    \hfill
    \includegraphics[width=0.24\columnwidth]{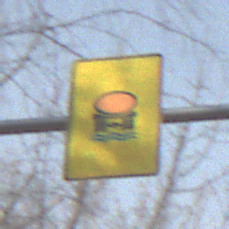}\\[3pt]

    \includegraphics[width=0.24\columnwidth]{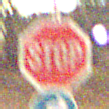}
    \hfill
    \includegraphics[width=0.24\columnwidth]{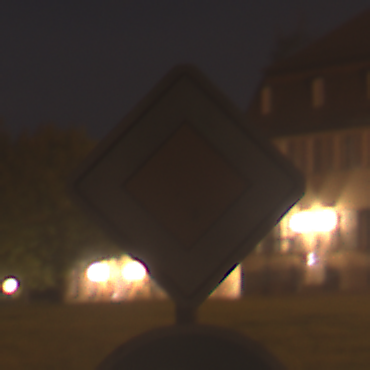}
    \hfill
    \includegraphics[width=0.24\columnwidth]{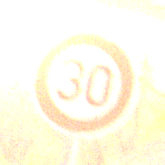}
    \hfill
    \includegraphics[width=0.24\columnwidth]{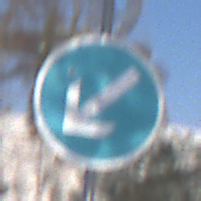}
    \caption{Example images of Synset Signset Germany including challenging conditions as, e.g., noisy, night, overexposed, or shadowed images (lower row).}
    \label{fig:example-images}
    \vspace{-9pt}
\end{figure}

\begin{figure}[H]
    \centering
    \begin{subfigure}[b]{0.239\columnwidth}
        \includegraphics[width=\columnwidth]{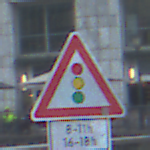} 
        
        \vspace{1mm}
        \includegraphics[width=\columnwidth]{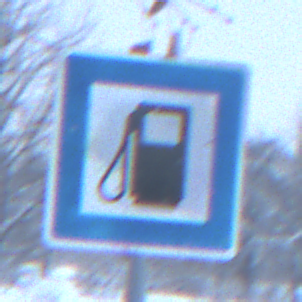}
        \caption{Cycles}
    \end{subfigure}
    \hfill
    \begin{subfigure}[b]{0.239\columnwidth}
        \includegraphics[width=\columnwidth]{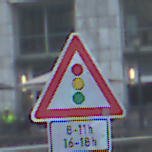}
        
        \vspace{1mm}
        \includegraphics[width=\columnwidth]{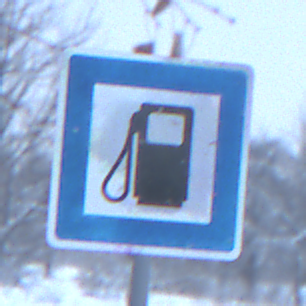}
        \caption{OGRE}
    \end{subfigure}
    \hfill
    \begin{subfigure}[b]{0.239\columnwidth}
        \includegraphics[width=\columnwidth]{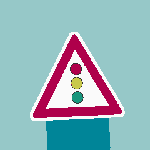}
        
        \vspace{1mm}
        \includegraphics[width=\columnwidth]{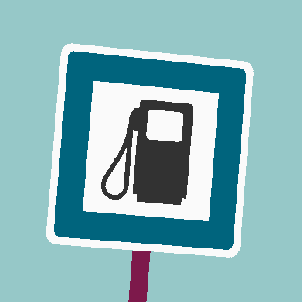}
        \caption{Segmentation}
    \end{subfigure}
    \hfill
    \begin{subfigure}[b]{0.239\columnwidth}
        \includegraphics[width=\columnwidth]{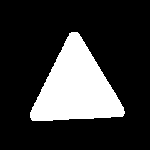}
        
        \vspace{1mm}
        \includegraphics[width=\columnwidth]{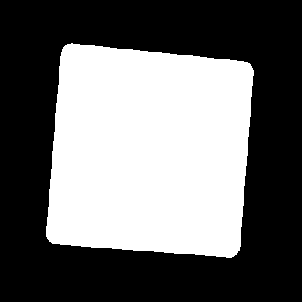}
        \caption{Mask}
    \end{subfigure}    
    \label{fig:example-triplets}
    \vspace{-9pt}
\end{figure}

In this context, the use of synthetic data has been considered for various roles: Most commonly, synthetic data can reduce costs and effort compared to real-world data acquisition; for example, \cite{Cordts2016Cityscapes} cites an average of 90 minutes for annotation and quality control of a single image of pixel-level segmentation within the Cityscapes dataset, whereas the same level of annotation can be extracted directly as ground truth from a simulation (e.g., \cite{johnson2016driving, gaidon2016virtual}). Here, the main focus is on the substitution of \emph{training data} through synthetic data, primarily because usually larger quantities of training data are used in the development of ML systems compared to during testing. Depending on the application, synthetic data are used as an extension to available real-world training datasets. Simulated images also provide a means of producing data for rare or dangerous scenarios that can hardly be collected under real-world conditions, which can benefit training as well as testing data.

\begin{table*}
    \caption{Overview of the most relevant publicly available traffic sign recognition datasets sorted by year of publication. For datasets of type rec (recognition), the number of traffic sign instances is equal to the number of samples.}
    \centering
    \small
    \begin{tabular}{rcccccccc}
        \toprule
        Dataset & Year & Type & \# Images & \# Samples & \# Classes & $\varnothing$ Samples/class & Real syn & Region \\
        \midrule
        MASTIF \cite{MASTIF} & 2009 & rec & 6\,428 & 6\,428 & 94 & 68.4 & real & Croatia \\
        MASTIF \cite{MASTIF} & 2010 & det \& rec & 3\,889 & 5\,215 & 86 & 60.6 & real & Croatia \\
        Stereopolis \cite{Stereopolis} & 2010 & det \& rec & 847 & 251 & 10 & 25.1 & real & France \\
        MASTIF \cite{MASTIF} & 2011 & det \& rec & 1\,013 & 1\,473 & 51 & 28.9 & real & Croatia \\
        STS (set 1\&2) \cite{larsson2011using} & 2011 & det \& rec & 3\,777 & 6\,652 & 19 & 350.1 & real & Sweden \\
        GTSRB \cite{GTSRB} & 2011 & rec & 51\,882 & 51\,882 & 43 & 1\,206.6 & real & Germany \\
        LISA \cite{LISAdataset} & 2012 & det \& rec & 6\,610 & 7\,855 & 49 & 160.3 & real & USA \\
        BTSC \cite{BelgiumTS} & 2013 & rec & 7\,125 & 7\,125 & 62 & 114.9 & real & Belgium \\
        TT100K \cite{TT100K} & 2016 & det \& rec & 100\,000 & 30\,000 & 221 & 135.7 & real & China \\ 
        CURE-TSR \cite{CURE-TSR} & 2017 & rec & 2\,206\,106 & 2\,206\,106 & 14 & 157\,579.0 & mixed & Belgium\\
        TSRD \cite{TSRD} & 2018 & rec & 6\,164 & 6\,164 & 58 & 106.3 & real & China \\
        European DS \cite{EuropeanDataset} & 2018 & rec & 82\,476 & 82\,476 & 164 & 502.9 & real & Europe \\
        DFG \cite{DFGTrafficSignDataset} & 2019 & det \& rec & 6\,957 & 17\,598 & 200 & 88.0 & real & Slovenia \\
        \small{fully annot.} MTSD \cite{MapillaryTrafficSignDataset} & 2020 & det \& rec & 52\,453 & 257\,541 & 400 & 643.9 & real & Global \\
        \small{part. annot.} MTSD \cite{MapillaryTrafficSignDataset} & 2020 & det \& rec & 53\,377 & 96\,613 & 400 & 241.5 & real & Global \\
        CATERED \cite{CATERED} & 2021 & rec & 94\,478 & 94\,478 & 43 & 2\,197.2 & syn & Germany \\
        \rowcolor{fraunhoferLightBlue}Synset Signset Ger. (ours) & 2024 & rec & 105\,500 & 105\,500 & 211 & 500.0 & syn & Germany \\
        \bottomrule
    \end{tabular}
    \label{tab:overwiew_traffic_sign_datasets}
\end{table*}

Beyond this, synthetic data can also provide an approach to dependable AI, by analyzing the performance of ML systems---and particularly their sensitivity to parameters---more systematically and quantitatively, specifically when used as \emph{testing data}. This is particularly important for determining, i.a., the robustness of AI systems. 
However, to what extent these benefits can be leveraged depends strongly on the degree of realism in the synthetic data. The more pronounced this ``sim-to-real'' gap is, the less reliable conclusions are, such as conclusions about the performance of a given ML system in the real world, within the intended operational design domain (ODD).

A particularly important regulation on requirements for the use of training and testing data for AI/ML applications is the European AI Act, proposed in 2021 and expected to become law in mid-2024, stating in the texts adopted in the March 2024 resolution ``Data sets for training, validation and testing, including the labels, should be relevant, sufficiently representative, and to the best extent possible free of errors and complete in view of the intended purpose of the system'' (with the clause ``to the best extent possible'' added compared to the 2021 proposition) \cite{aia}. In this context, the quantitative comparison between domain gaps for a choice of real vs. synthetic data sources is expected to gain highly practical relevance, particularly for high-risk applications identified within the AI Act, such as ``AI systems intended to be used as safety components in the management and operation of road traffic'' \cite[Annex III]{aia-annex}.

In this context, the task of traffic sign recognition plays multiple roles that extend beyond the immediate purpose of classifying traffic sign images into their legal categories and semantics, for example for driver assistance systems, automated driving, and mapping. Traffic sign recognition is an extensively researched topic across a wide range of methods \cite{BelgiumTS, de1997road, de2003traffic, bahlmann2005system, maldonado2007road, fu2010survey}, spanning a range from completely analytic approaches over classical ML with tailored models and features up to modern deep learning. In this domain, \cite{ciregan2012multi} is commonly cited as the first instance where a machine learning approach outperformed humans on a complex computer vision task, with the presented multi-column deep neural networks (MCDNN) achieving half the error rate of humans on the \emph{German Traffic Sign Recognition Benchmark} (GTSRB) dataset \cite{GTSRB_Stallkamp2012}. At the same time, with new traffic signs constantly being released and coverage of existing signs in datasets still limited for a distinction of less common classes, the demand for both training and testing data still persists. This connection between a large body of recognition methods with still highly topical applications on a task that provides a relatively controlled scope motivates the choice of traffic sign recognition for an analysis of synthetic data for training and reliability assessment via XAI and robustness checks, and the generation of a novel simulated dataset.

\section{STATE OF THE ART}\label{sec:sota}

\begin{figure*}
    \begin{subfigure}[b]{0.094\textwidth}
    \includegraphics[width=\columnwidth, trim=2pt 0 0 0, clip]{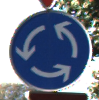}
    \includegraphics[width=\columnwidth, trim=1.8pt 0 0 0, clip]{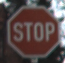}
    \includegraphics[width=\columnwidth, trim=2pt 0 0 0, clip]{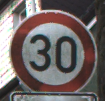}
    \caption{\cite{GTSRB}\\GTSRB}
    \end{subfigure}
    \hfill
    \begin{subfigure}[b]{0.094\textwidth}
    \includegraphics[width=\columnwidth, trim=3pt 0 0 0, clip]{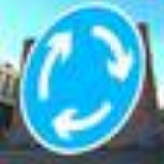}
    \includegraphics[width=\columnwidth]{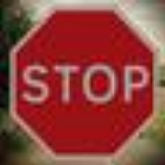}
    \includegraphics[width=\columnwidth]{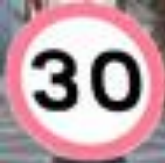}
    \caption{\cite{SynTrafSignRec_SimpleRandPlacement}\\Augmentation}
    \end{subfigure}
    \hfill
    \begin{subfigure}[b]{0.094\textwidth}
    \includegraphics[width=\columnwidth]{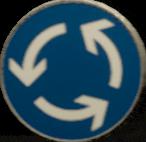}
    \includegraphics[width=\columnwidth]{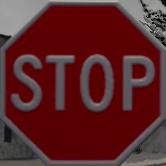}
    \includegraphics[width=\columnwidth, trim=3pt 0 0 0, clip]{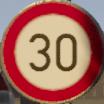}
    \caption{\cite{CATERED}\\CATERED}
    \end{subfigure}
    \hfill
    \begin{subfigure}[b]{0.094\textwidth}
    \includegraphics[width=\columnwidth, trim=0 0 2pt 0, clip]{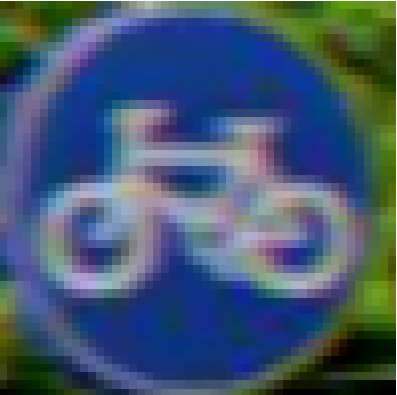}
    \includegraphics[width=\columnwidth]{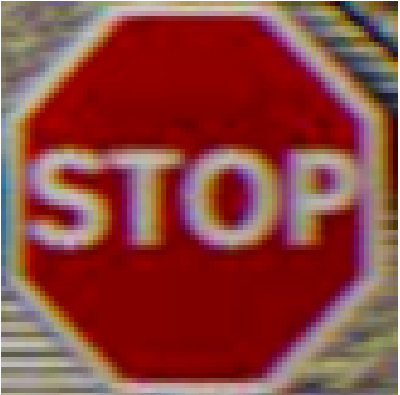}
    \includegraphics[width=\columnwidth]{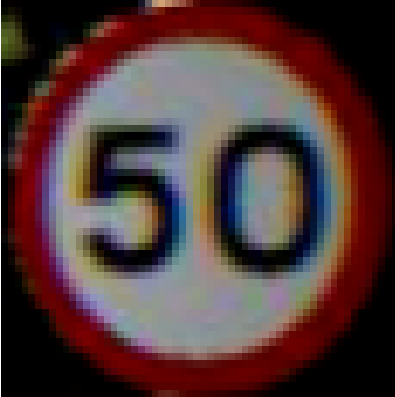}
    \caption{\cite{CURE-TSR}\\CURE-TSR}
    \end{subfigure}
    \hfill
    \begin{subfigure}[b]{0.094\textwidth}
    \includegraphics[width=\columnwidth,trim=36 36 36 36, clip]{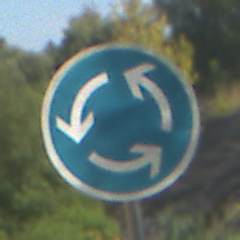}
    \includegraphics[width=\columnwidth,trim=33 33 33 33, clip]{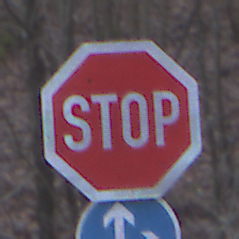}
    \includegraphics[width=\columnwidth,trim=40 32 34 42, clip]{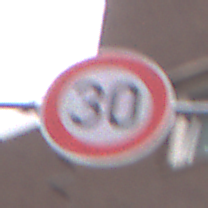}
    \caption{\textcolor{fraunhoferBlue}{Ours\\Synset Signset}}
    \end{subfigure}
    \hfill
    \begin{subfigure}[b]{0.094\textwidth}
    \includegraphics[width=\columnwidth,trim=10.6 10 10 10, clip]{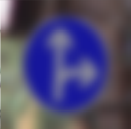}
    \includegraphics[width=\columnwidth,trim=17.6 17 17 17, clip]{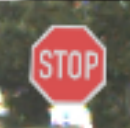}
    \includegraphics[width=\columnwidth,trim=16.6 16 16 16, clip]{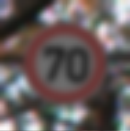}
    \caption{\cite{SynTrafSignRec_GANandPlacement}\\Augmentation}
    \end{subfigure}
    \hfill
    \begin{subfigure}[b]{0.094\textwidth}
    \includegraphics[width=\columnwidth,trim=16 14 16 18, clip]{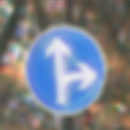}
    \includegraphics[width=\columnwidth,trim=10 8 10 12, clip]{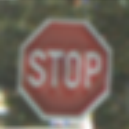}
    \includegraphics[width=\columnwidth,trim=11 9 11 13, clip]{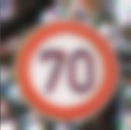}
    \caption{\cite{SynTrafSignRec_GANandPlacement}\\GAN}
    \end{subfigure}
    \hfill
    \begin{subfigure}[b]{0.094\textwidth}
    \includegraphics[width=\columnwidth,trim=1.4pt 0 1.4pt 0, clip]{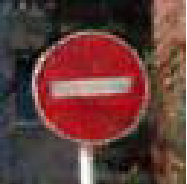}
    \includegraphics[width=\columnwidth,trim=1.4pt 0 1.4pt 0, clip]{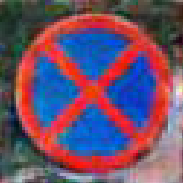}
    \includegraphics[width=\columnwidth,trim=1.2pt 0 1.2pt 0, clip]{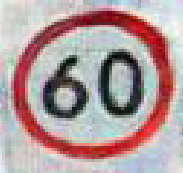}
    \caption{\cite{SynTrafSignRec_GanApproach}\\DCGAN}
    \end{subfigure}
    \hfill
    \begin{subfigure}[b]{0.094\textwidth}
    \includegraphics[width=\columnwidth, trim=0 0 0 5pt, clip]{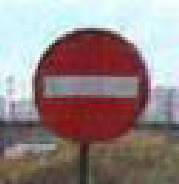}
    \includegraphics[width=\columnwidth]{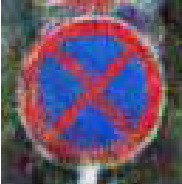}
    \includegraphics[width=\columnwidth]{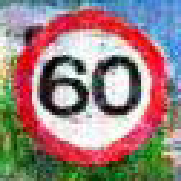}
    \caption{\cite{SynTrafSignRec_GanApproach}\\LSGAN}
    \end{subfigure}
    \hfill
    \begin{subfigure}[b]{0.094\textwidth}
    \includegraphics[width=\columnwidth]{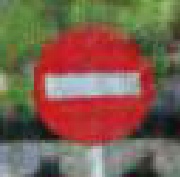}
    \includegraphics[width=\columnwidth]{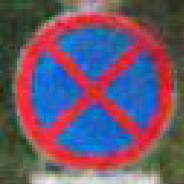}
    \includegraphics[width=\columnwidth]{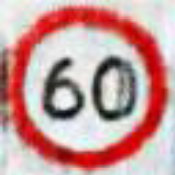}
    \caption{\cite{SynTrafSignRec_GanApproach}\\WGAN}
    \end{subfigure}
    
    \caption{Comparison of real images represented by GTSRB (left) and state-of-the-art image synthetization methods for traffic sign recognition. For achieving a better comparability we cropped the images to a similar area if necessary. The DCGAN, LSGAN, and WGAN samples stemming from \cite{SynTrafSignRec_GanApproach} result from training 200 epochs respectively.}
    \label{fig:stateOfTheArtImageComparison}
\end{figure*}

\subsection{Publicly available Traffic Sign Recognition Datasets}

\cref{tab:overwiew_traffic_sign_datasets} provides an overview of the most relevant publicly available datasets for the task of traffic sign recognition. Datasets of type ``recognition'' (``rec'') already contain images cropped to approximately the sign size, while datasets of type ``detection and recognition'' (``det \& rec'') show entire street scenes that must be cropped using specified bounding boxes. Most of the datasets are only valid for certain countries. The best known among the listed datasets are GTSRB~\cite{GTSRB},  \emph{Tsinghua-Tencent 100K} (TT100K)~\cite{TT100K}, the \emph{European Dataset}~\cite{EuropeanDataset} (which includes, i.a., \cite{MASTIF,Stereopolis,larsson2011using,GTSRB,RUG}), and the \emph{Mapillary Traffic Sign Dataset} (MTSD)~\cite{MapillaryTrafficSignDataset}.

\subsection{Usage of Synthetic Data}

The comparison of synthetic data with real data on fine-grained classification was presented in \cite{anne-icra} on the example of the \emph{Synset Boulevard} dataset for the task of vehicle make and model recognition (VMMR). This study found synthetic data to be generally capable of achieving performance comparable to training on the real-world CompCars dataset \cite{CompCarsDataset} (cf. \cite{anne-icra} also for a broader overview of synthetic data use in mobility). In the specific field of traffic sign recognition, many authors use synthetic data to increase the volume of training and/or test data, especially for rare classes. Commonly applied approaches for such synthetic data generation are:

\subsubsection{Image Augmentation} 
In general, image augmentation methods (e.g., \cite{SynTrafSignRec_SimpleRandPlacement}) implement the following steps: They collect traffic sign templates, apply an affine transformation on them for diversifying the sign rotations and scales, vary the sign hue and/or saturation values by possibly including the background image properties or adapting the background patches, combine templates and backgrounds, and---if applicable---deploy post processing  such as blur. In \cite{SynTrafSignRec_RandomPlacementApproachWithDR}, domain randomization is used additionally, while  \cite{SynTrafSignRec_RandomPlacementApproach} expands this procedure by randomly inserting computer generated traffic signs (for one experiment also with GAN-generated textures) to background images. The authors show that image augmentation approaches are able to expand real-world datasets in a targeted manner, but there are still disadvantages, e.g., that DNNs could overfit on domain differences or insertion edges, and that signs without dirt or wear artifacts oversimplify the classification task.  

\subsubsection{Simulations} 
For the creation of the CATERED dataset \cite{CATERED} the Carla Simulator\footnote{\href{https://carla.org/}{carla.org}} was utilized. The authors of CURE-TSR \cite{CURE-TSR} expanded their dataset by adding simulated images generated by using the Unreal Engine 4\footnote{\href{https://www.unrealengine.com/}{unrealengine.com}}. It is also conceivable to employ computer games, as already practiced for automotive datasets for object recognition \cite{gta5objectDetection} or semantic segmentation \cite{gta5SemanticSegmentation}. With this approach, it is important to ensure that the simulation environment offers sufficient variance and that the traffic signs are not oversimplified in order to achieve an adequate degree of realism, so that the sim-to-real gap is kept as small as possible and that unrealistic overfitting is prevented.

\subsubsection{Generative Adversarial Networks (GANs)} 
Other approaches, such as \cite{SynTrafSignRec_GANandPlacement} and \cite{SynTrafSignRec_GanApproach}, use generative adversarial networks (GANs) to generate additional training data leading to an improvement of classification results. GANs are able to increase the degree of realism compared to the previous described approaches. However, referring to \cite{SynTrafSignRec_GANandPlacement}, applying the geometric transformation through the GAN is challenging. This is why the authors therefore implemented traditional methods and used the GANs only for synthesizing the visual appearance. This can also be observed in the results of \cite{SynTrafSignRec_GanApproach}, as the geometric shapes of the signs are partly imprecise. Furthermore, this approach relies on training data, which are difficult to collect for rarely occurring traffic signs.

\cref{fig:stateOfTheArtImageComparison} compares images of all the approaches mentioned.

\section{SYSTEMATIC SYNTHETIZATION:\linebreak DATASET GENERATION AND COMPOSITION}\label{sec:generation}

\begin{figure*}
    \centering
    \includegraphics{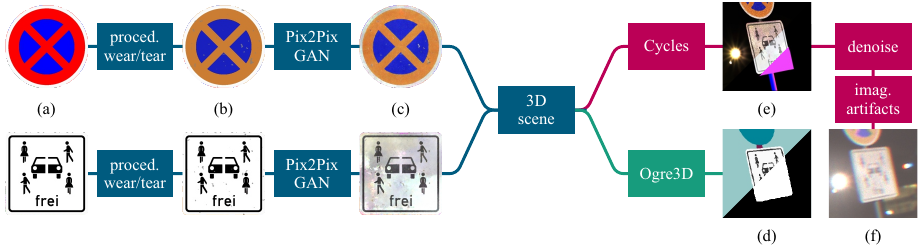}
    \caption{Overview of the generation pipeline built in OCTANE. Ideal images (a) are procedurally converted to template images (b) defining color degradation and wear/tear masks. A GAN trained on worn traffic signs converts these to diffuse textures (c) that are combined into a 3D scene for physically-based rendering. Segmentation and mask images (d) are rendered using OGRE, while Cycles is used for geometric raytracing of HDR raw image, albedo, and normal image (e). The latter are used to denoise the raytracing samples. Based on this, imaging artifacts are computed on the 2D image data (f).}
    \label{fig:pipeline}
\end{figure*}

The dataset was generated through a systematic synthetization approach shown in \cref{fig:pipeline}, distinguishing between factors that require learning distributions from training data and factors that can be modeled analytically. The approach aims to support explainable datasets, where each aspect in the pipeline is associated with a model that is self-contained and specified as clearly as possible w.r.t. assumptions and characteristics.

\subsection{Texture and Defect Generation}\label{sec:texture-generation}

The visual appearance of traffic signs is prominently affected by deterioration of the sign surface, through wear, tear, vandalism, or fading of colors. These effects are complex and difficult to model analytically; hence, a primarily data-driven approach was chosen to introduce realistic defects into textures.

The main goal of the particular approach is to achieve a realistic distribution of defects at variable intensities across a potentially unlimited set of traffic signs. Hence a model was designed that can be trained on acquired data but does not depend on particular sign shapes and can use medium-level annotations to selectively apply defects.

\subsubsection{GAN-based synthesis from template images}

We apply a Pix2Pix-based generative adversarial model (GAN) \cite{pix2pix_isola2017image} without the central $1\times 1 \times 512$ bottleneck to achieve a fully convolutional layout that can adapt to given input / output dimensions. With this layout, we train the GAN to convert texture patches of arbitrary size at a fixed spatial resolution of 8\,px/cm, containing template images of arbitrary shapes, into the equivalent texture patch with defects. Through this, the GAN can generalize towards new physical sign sizes and new shapes; however, possible correlations between sign type (rather than visual shape) and damages (e.g., specific dirt on wild animals crossing signs) will be largely eliminated.

The GAN is trained on 200+ worn traffic signs where the color/dirt templates were extracted through classical image processing (cf. \cref{fig:gan_train_in}--\subref{fig:gan_train_out}). Color templates support black, white, and saturated colors. Gray spots annotate dirt and scratches---hence, gray is not supported as a sign color, limiting some existing variants of German traffic signs. Retroreflector patterns are excluded and retroreflection is not simulated. Pairs of mask (input) and raw (output) images are generated by randomly cropping and rotating the original images to patches of $256^2\times 3$ and randomly shuffling the RGB channels to increase the color variation, since yellow, green, cyan, and purple hues are underrepresented or not represented at all in the original dataset.

The output textures are used exclusively as the diffuse component in the PBR (physically-based rendering, cf.~\cref{sec:rendering}) surfaces.

\subsubsection{Generation of template images from sign shapes}

\begin{figure}
    \begin{subfigure}{\columnwidth}
    \includegraphics{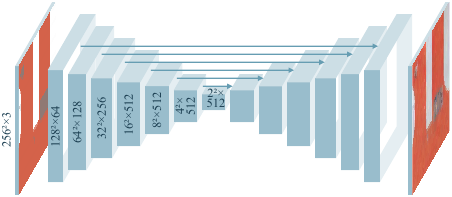}
    \caption{Fully-convolutional Pix2Pix structure with $2^2\times512$ central bottleneck.}
    \label{fig:gan_architecture}
    \end{subfigure}\\

    \begin{subfigure}{0.235\columnwidth}
    \includegraphics[width=\columnwidth]{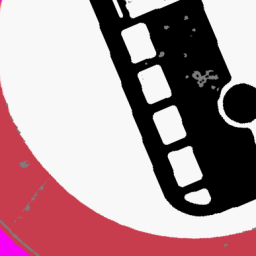}
    \caption{train. input sample ($256^2{\times}3$)}
    \label{fig:gan_train_in}
    \end{subfigure}
    \hfill
    \begin{subfigure}{0.235\columnwidth}
    \includegraphics[width=\columnwidth]{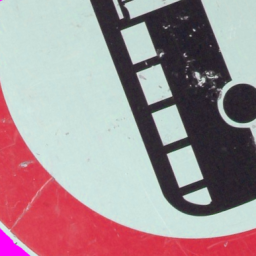}
    \caption{train. output sample ($256^2{\times}3$)}
    \label{fig:gan_train_out}
    \end{subfigure}
    \hfill
    \begin{subfigure}{0.235\columnwidth}
    \includegraphics[width=\columnwidth]{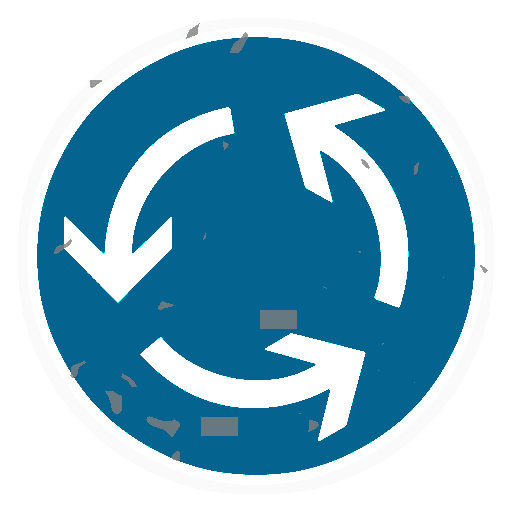}
    \caption{infer. input sample ($512^2{\times}3$)}
    \label{fig:gan_inf_in}
    \end{subfigure}
    \hfill
    \begin{subfigure}{0.235\columnwidth}
    \includegraphics[width=\columnwidth]{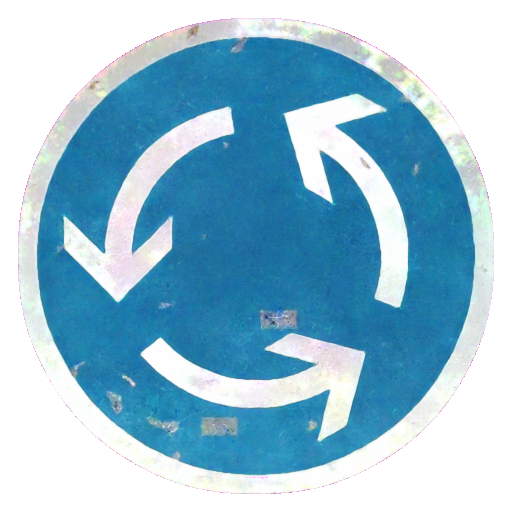}
    \caption{infer. output sample ($512^2{\times}3$)}
    \label{fig:gan_inf_out}
    \end{subfigure}
    \label{fig:gan}
\caption{GAN setup for defect synthetization.}
\end{figure}

The template images that are used as input to the GAN (\cref{fig:gan_inf_in}) are generated from the Wikipedia overview of German traffic signs\footnote{\href{https://de.wikipedia.org/wiki/Bildtafel_der_Verkehrszeichen_in_der_Bundesrepublik_Deutschland_seit_2017}{de.wikipedia.edia}org/wiki/Bildtafel\textunderscore{}der\textunderscore{}Verkehrszeichen\textunderscore{}in\textunderscore{}der\newline\textunderscore{}Bundesrepublik\textunderscore{}Deutschland\textunderscore{}seit\textunderscore{}2017 \label{fn:wikipedia}}. The signs are separated into black, white, red, orange, yellow, green, and blue components. Each color component is faded stochastically and homogeneously across each sign based on the stochastic distribution of the real sign samples. Subsequently, a gray dirt mask, procedurally generated through a noise process, is overlaid, combining arbitrary shapes and rectangular shapes, the latter representing worn stickers that occur frequently in the real dataset.

\subsection{Scene Variation}

The scene variation and rendering of Synset Signset Germany is performed by the Fraunhofer simulation platform OCTANE\footnote{\href{https://www.octane.org/}{octane.org}}, written in C++ and following a plugin-based architecture. The following scene variations are applied:

\subsubsection{Traffic sign material} 
Each traffic sign instance is assigned a unique texture generated as described in \cref{sec:texture-generation}. In addition, the roughness component in the PBR surface is uniformly varied in the interval between $0.2$ and $0.4$, the specular component between $0.3$ and $0.5$.

\subsubsection{Traffic sign pole}
We distinguish traffic sign classes into those that are to be exclusively featured on vertical poles, and those that can occur both on vertical and on horizontal poles.
In the second case, a horizontal pole is uniformly chosen with a probability of $0.3$, a vertical pole otherwise. The pole diameter varies in the vertical case between 8\,cm and 12\,cm, in horizontal case between 8\,cm and 20\,cm. 
The roughness (between $0.4$ and $0.6$), and the diffuse color ($R{=}G{=}B \in [0.25,0.4]$) of the poles' PBR surface are also varied.

\subsubsection{Number of signs per pole} 
For each traffic sign in our dataset, we manually labeled permissible possible upper and lower signs by taking the German traffic code / regulation  StVO\footnote{\href{https://www.stvo2go.de/verkehrszeichen-wissensnetz/}{stvo2go.de/verkehrszeichen-wissensnetz}} (Straßenverkehrs-Ordnung) and real-world examples into account. Thereby, we have not only considered the 211 traffic signs contained in Synset Signset Germany, but also 135 additional supplementary traffic signs. Additional traffic signs are only added to vertical poles with a probability of $0.5$ to increase the dataset's level of difficulty.

\subsubsection{Camera orientation}
We choose the camera orientation as follows: In case of a vertical pole, $\text{roll}\sim\mathcal{N}(0.0^\circ,2.0^\circ)$, $\text{pitch}\sim\mathcal{N}(5.0^\circ,10.0^\circ)$, and $\text{yaw}\sim\mathcal{N}(0.0^\circ,21.0^\circ)$. For horizontal poles, we define a smaller yaw orientation range but higher pitch mean, namely $\text{roll}\sim\mathcal{N}(0.0^\circ,2.0^\circ)$, $\text{pitch}\sim\mathcal{N}(30.0^\circ,10.0^\circ)$, and $\text{yaw}\sim\mathcal{N}(0.0^\circ,16.0^\circ)$. The camera is positioned so that it is centered on the traffic sign. 

\subsubsection{Environment} To modulate the environment and lighting, our approach uses image-based lighting (IBL) based on 327 uniformly sampled environment maps collected from Polyhaven\footnote{\href{https://polyhaven.com/}{polyhaven.com}}. Moreover, their azimuth is also varied uniformly.

\subsubsection{Occlusion object} To cast shadows, a 3D tree object is randomly placed in the scene for $\nicefrac{3}{4}$ of the images. Whether the shadow is visible on the sign also depends on the random position of the sun.

\subsection{Optical Simulation / Rendering Pipeline}\label{sec:rendering}

The optical simulation in OCTANE follows the general framework of physically-based rendering (PBR) \cite{pharr2023physically} which provides approximately consistent models for light transport in the scene using a common set of properties. This enables the exchange of ``solvers'' for image generation, for which OCTANE currently supports the rasterization-based engine OGRE\footnote{\href{https://ogre3d.org/}{ogre3d.org}} as well as the path tracing engine Cycles\footnote{\href{https://cycles-renderer.org/}{cycles-renderer.org}} from the Blender project. We provide all 105\,500 Synset Signset Germany images rendered by Cycles and OGRE respectively.
For the XAI and robustness analysis the rendering was performed by OGRE to be able to test more configurations due to the reduced amount of render time. 
The segmentation masks and mask images where created by using OGRE.

As an approximation for the complex light transport in the scene, the modeling separates into an idealized geometric light tracing in the scene purely based on ray / surface interactions, and the computation of convolutional effects and degradations based on the resulting high dynamic range raster images.

Thus, subsequent computations after the geometric rendering include the following:
\begin{itemize}
    \item Stochastic errors in automatic exposure control (AEC) and white balance (WB) as presented in \cite{anne-icra}.
    \item Simulation of the \emph{point spread function} (PSF) based on a Tamron M112FM35 35~mm lens to represent focusing, lens optics, and diffraction through a mixture-of-Gaussian model as presented in \cite{anne-icra}.
    \item Simulation of \emph{lens flares} for visible light sources and lighting-dependent \emph{noise}, each as presented in \cite{anne-icra}.
    \item Simulation of \emph{motion blur} and \emph{chromatic aberration} through linear convolution kernels in arbitrary direction with uniformly distributed length $\sim \mathcal U(0\,\text{px}, 10\,\text{px})$.
    \item Simulation of \emph{digital image sharpening} effects using unsharp masking.
    \item Addition of artifacts from Bayer BGGR bilinear \emph{demosaicing} as in \cite{anne-icra}.
\end{itemize}

For all simulated effects and artifacts, the stochastically selected parameter values are given per individual image in the dataset.

\subsection{Dataset Statistics}

\begin{figure*}
    \centering
    \includegraphics[width=\textwidth]{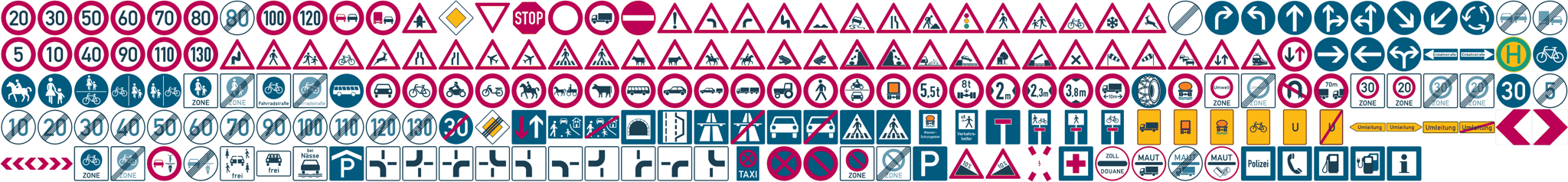}
    \caption{Overview of the signs in Synset Signset Germany. The first row of 43 signs corresponds to the classes in GTSRB. Sign shapes are based on the Wikipedia overview of German traffic signs from 2017 onwards (cf. \cref{fn:wikipedia}).}
    \label{fig:signlist}
    \vspace{-9pt}
\end{figure*}

Our resulting dataset \emph{Synset Signset Germany} contains 211 traffic sign classes depicted in \cref{fig:signlist}. The dataset is balanced with 500 images per class, resulting in a total of 105\,500 images. Thereby, the traffic sign classes can be grouped as follows:\\

{\small
\noindent\begin{tabular}{@{}rlcrl}
17 & Speed limit signs & & 45 & Danger signs\\
33 & Other prohibitory signs & & 21 & Derestriction signs \\
12 & Stop, wait, and parking signs & & 28 & Information signs \\
13 & Driving lane control signs & & 4 & Priority signs \\
13 & Special zones and way signs & & 4 & Highway signs \\
13 & Additional road signs & & 8 & Other signs
\end{tabular}
}\\

The image resolutions in our dataset vary between the maximum resolution of 389\,$\times$\,389 pixels and the minimum resolution of 22\,$\times$\,22 pixels.

\section{SYNTHETIC DATA AS TRAINING DATA}\label{sec:training}

To determine the degree of realism in the synthetic data, we evaluate our dataset in comparison to the GTSRB dataset~\cite{GTSRB} based on the subset of the first 43 classes in Synset Signset that overlaps with GTSRB. Additionally, we utilize the CATERED~\cite{CATERED,CATERED_profPaper} dataset for training and evaluation as a synthetic reference dataset. For all experiments, we employ a ConvNeXt-Small~\cite{ConvNeXt} network with similar settings as Sielemann et al.~\cite{anne-icra}. We only refrain from applying random flip augmentation since the orientation of some signs are a distinguishing feature, and we utilize different learning rates. For the evaluation on CATERED, we additionally remove the center crop and instead directly resize to 224x224 since the images in CATERED are already tightly cropped.

Regarding the learning rates, we train models with learning rates of $10^{-4}$, $10^{-3}$, and $10^{-2}$ and choose the learning rate with the best in-domain result for each of the datasets. This is done to choose the learning rate which is most appropriate for training on each dataset while reducing the risk of overfitting in the cross-domain evaluations. We apply an 80--20 split to Signset for extracting a training and a validation set while utilizing the official train--validation splits for the other datasets. The results as measured with top-1 accuracy are shown in \Cref{tab:resultsclassification}. They show an accuracy above 80\,\% for the evaluation on all three datasets when training on Synset Signset Germany. For the evaluation in the cross-dataset scenarios, the scores are just closely behind the in-domain trainings, only lacking 1.2 percentage points when evaluating on the real-world GTSRB dataset, while the evaluation on Signset shows the highest score by a large margin. This highlights the usefulness of our dataset for training classification models. Moreover, the large margin between training on Signset compared to training on one of the other datastes for an evaluation on Signset indicates the challenge of the training dataset, and thus, a high usefulness for evaluation purposes considering the saturation of possible improvements on the GTSRB. It additionally provides a significantly higher value due to the inclusion of a total of 211 instead of 43 classes, with a model trained on all classes still achieving a score of 99.6\,\% for both full Signset dataset versions, Cycles and OGRE.

\begin{table}
    \centering
    \small
        \caption{Top-1 accuracy of each combination of training and testing on the considered datasets. The results indicate the high effectiveness of Signset for training as well as for evaluation purposes.}
    \begin{tabular}{ccccc}
        \toprule
           Eval. \rlap{$\blacktriangleright$}& Signset & Signset & \multirow{2}{*}{GTSRB} & \multirow{2}{*}{CATERED} \vspace{-1mm} \\
           Train.$\blacktriangledown$ & Cycles & OGRE &  & \\
        \midrule
        Signset\vspace{-1mm} & \multirow{2}{*}{99.5\%} & \multirow{2}{*}{99.4\%} & \multirow{2}{*}{98.3\%} & \multirow{2}{*}{84.4\%} \\
        Cycles & & & & \\
        Signset\vspace{-1mm}& \multirow{2}{*}{99.6\%} & \multirow{2}{*}{99.6\%} & \multirow{2}{*}{98.2\%} & \multirow{2}{*}{84.6\%} \\
        OGRE & & & & \\
        GTSRB & 89.4\% & 87.4\% & 99.9\% & 77.1\% \\
        CATERED & 50.0\% & 48.6\% & 76.4\% & 86.1\% \\
        \bottomrule
    \end{tabular}
    \label{tab:resultsclassification}
\end{table}


\section{SYNTHETIC DATA FOR\linebreak XAI AND ROBUSTNESS ANALYSIS}\label{sec:xai}

Synthetic data can also play a relevant role in the investigation of ML models in terms of robustness (stability of the model prediction performance w.r.t. input perturbations) and explainability, which attempts to explain model decisions in order to increase their comprehensibility. 
Both of these aspects are important in assessing the reliability of a trained AI/ML system, especially if that system is to be used in a safety-relevant context.

\begin{figure}
    \includegraphics[width=\columnwidth]{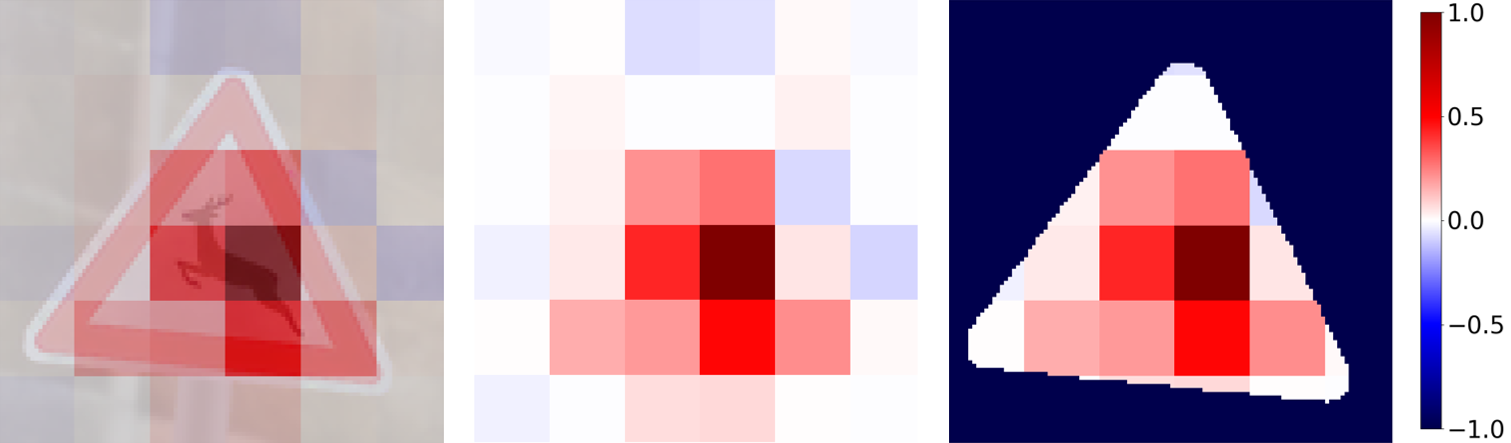}
    \caption{Example image with its corresponding FA map overlaid on the left. The center image shows the pure FA map, and the right image shows only the features (i.e., the image pixels) with increased attribution value that make up the traffic sign.}
    \label{fig:hm_example}
    \vspace{-10pt}
\end{figure}

In this regard, we use the Synset Signset Germany synthesis pipeline to evaluate specific explainability and global robustness measures. By introducing certain parametric perturbations, we can ascertain the explanation quality and robustness level for arbitrary parameters.
Explanation quality is quantified via the \emph{pixel ratio}, under the premise that a ``good'' explanation should predominantly highlight the object of interest rather than something in the background. In the experiments described, explanations are generated using local saliency methods that yield \emph{feature attribution} (FA) maps per image. These maps, when combined with binary mask images (cf. \cref{fig:example-triplets}, right), enable us to assess the amount of attributed features (i.e., pixels with a positive attribution value) that belong to the traffic sign. Specifically, the pixel ratio is defined as the proportion of positively attributed pixels---weighted by their attribution value---constituting the traffic sign relative to those in the entire image. This definition is illustrated in \cref{fig:hm_example}: The pixel ratio is the ratio of the positive FA in the right image to the positive FA in the center image. The higher its value, the better the explanation. The FA was computed using the KernelSHAP method from the Captum library\footnote{\href{https://captum.ai/}{captum.ai}}.

Global robustness is evaluated through the method described in \cite{GloRo}. The evaluation is based on a sequence of hypothesis tests certifying a specific level of global robustness given a required confidence level. In contrast to the aforementioned method, instead of comparing the model prediction for the original input with the prediction for the perturbed input, here we compared the prediction for the original input with the ground-truth label. In that sense prediction performance for different perturbation intensities is used to assess robustness.

\begin{figure}
    \includegraphics[width=\columnwidth]{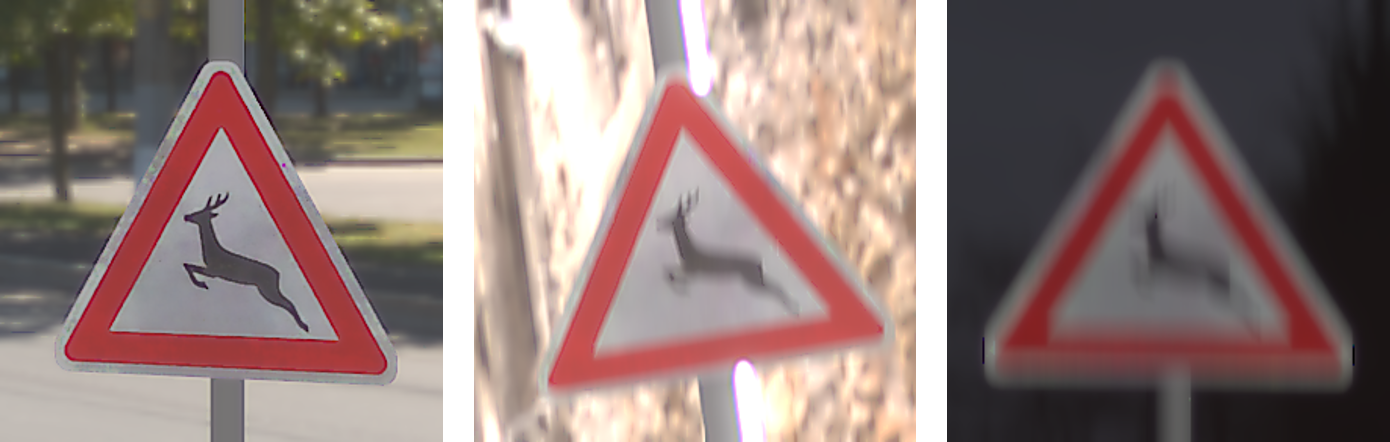}
    \caption{Example images for the three different datasets. On the left is an image with no motion blur, in the center an image with mid-level motion blur, and on the right an image with high-level motion blur.}
    \label{fig:mb_example}
\end{figure}

To demonstrate the approach, we consider a ResNet-18 \cite{Resnet} trained on the GTSRB dataset \cite{GTSRB}. However, the procedure works for any model. We focus on one specific perturbation, namely motion blur, and vary its intensity parameter in order to create three distinct traffic sign datasets for our experiments. These datasets correspond to none, mid-level, and high-level motion blur, as depicted in \cref{fig:mb_example}. The results of the experiments are reported in \cref{tab:xai_rob_results}.

\begin{table}
    \caption{Results of the XAI and robustness analysis on three different datasets with images of varying motion blur intensity.}
    \centering
    \small
    \begin{tabular}{rcc}
        \toprule
        Dataset & Pixel Ratio & Global Robustness \\
        \midrule
        No motion blur & 0.74 & 0.9 \\
        Mid-level motion blur & 0.63 & 0.88 \\
        High-level motion blur & 0.58 & 0.84 \\
        \bottomrule
    \end{tabular}
    \label{tab:xai_rob_results}
\end{table}

As expected, prediction performance as well as the quality of explanations (i.e., the pixel ratio) decrease with increasing motion blur intensity. From this it can be concluded that the model's performance diminishes when processing images with increasing levels of motion blur.

Overall, the benefit of Synset Signset Germany for XAI and robustness analysis lies in the ability to obtain quantitative measures for specific desired perturbation intensities. Furthermore, the inverse problem of finding intensity parameters to a required level of robustness and explanation quality can be addressed.

\section{CONCLUSION AND OUTLOOK}\label{sec:conclusion}

We have presented the Synset Signset Germany dataset, a synthetic dataset for the task of traffic sign recognition, containing a total of 105\,500 images of 211 different German traffic sign classes, including comparatively rare and very recent traffic signs. A subset of 43 classes in the dataset aims to represent a ``synthetic twin'' of the GTSRB dataset~\cite{GTSRB} with similar imaging parameters.

For each sign, detailed, stochastically chosen synthetization parameters are provided, along with additional binary mask and a segmentation mask label images. This is intended to support both the use of the dataset to understand machine learning effects on real data due to known ``ground truth'' parameters in the simulated images, and to understand the impact of different simulation methods on dataset quality.

Through this, the resulting dataset is among the largest and most diverse datasets for traffic sign recognition and, to the best of our knowledge, one of the first publicly available large-scale synthetic datasets for this task.

Our implemented synthesis pipeline proved to combine the advantages of data-driven and analytical modeling. Compared to the purely analytically simulated CATERED dataset, Synset Signset Germany achieves an approximately $20\,\%$ better top-1 accuracy, which, together with the high cross-dataset scores, indicates a good generalization ability probably due to the increased level of realism resulting from the GAN generated textures.


\subsection*{Outlook}

One of the main advantages in the use of synthetic data generation is its scalability towards further applications. Hence, based on the work, an important next step is to abandon the current limitation on German traffic signs and provide extensions towards international traffic signs.

A widely acknowledged limitation of synthetic data, in turn, is the sim-to-real domain gap. While the practical experiments indicate that the domain gap is sufficiently low for practical applications as training data, and that the data/metadata composition is well suited XAI, the requirements for the use as testing data are considerably higher. Here, the dataset not only has to cover the target domain sufficiently to train adequate generalization capabilities, but instead must also enable the quantitative performance estimation of trained models by relating effects on synthetic data to those on (yet unseen) real-world data. While the extensive annotation provided with the dataset is expected to support research in this area, i.e., by conducting XAI and robustness analyses for the remaining parameters, there is still considerable demand for future research.

The GAN-based defect synthesis so far uses only a very simple concept that does not distinguish between types of dirt and damage/wear and lacks representation of features such as gray sign areas and retroreflectors. Future work should improve on these limitations.

Furthermore, the number of occlusion objects should be increased in order to achieve more complex and diverse shadow casts and occlusions of the traffic signs. If the size of the synthetic dataset is to be significantly increased, it would be advisable to collect more environment maps to gain a higher data variance.

Eventually, the choice of recognition models in this paper is limited to few deep learning models within the state of the art. A more extensive analysis including more diverse models and potentially also some classical, non-deep learning approaches would substantiate the findings and extend the understanding of the applicability of the dataset.



\section*{ACKNOWLEDGMENT}

We would like to thank the civil engineering dept. of the city of Karlsruhe, Germany (Tiefbauamt Karlsruhe) for their kind support in creating the dataset of worn traffic signs.




\bibliographystyle{IEEEtran}
\bibliography{IEEEabrv,bibliography}

\end{document}

%% file: includes.tex
\usepackage{amsmath}
\usepackage{amssymb}

\usepackage[table]{xcolor}

\usepackage{tikz}
\usepackage{array}
\usepackage{framed, color}
\usepackage[normalem]{ulem}
\usepackage{cancel}
\usepackage{overpic}
\usepackage[footnotesize]{caption}
\usepackage[font+=footnotesize]{subcaption}
\usepackage[nodisplayskipstretch]{setspace}
\usepackage{txfonts}
\let\mathbb=\varmathbb

\usepackage{url}
\usepackage{upgreek}
\usepackage{float}
\usepackage{booktabs}
\usepackage{courier}
\usepackage[hidelinks]{hyperref}
\usepackage{nicefrac}
\usepackage{xcolor}
\usepackage{mdframed}
\usepackage{cleveref}
\usepackage{aveasmargins}

\usepackage{multirow}

\crefname{figure}{Fig.}{Figs.}
\Crefname{figure}{Fig.}{Figs.}

\crefname{table}{Tab.}{Tabs.}
\Crefname{table}{Tab.}{Tabs.}

\crefname{equation}{Eq.}{Eqs.}
\Crefname{equation}{Eq.}{Eqs.}

\crefname{section}{Sec.}{Secs.}
\Crefname{section}{Sec.}{Secs.}

\crefname{subsection}{Sec.}{Secs.}
\Crefname{subsection}{Sec.}{Secs.}